\documentclass{article}
\usepackage{ijcai19}

\usepackage{times}
\usepackage{soul}
\usepackage{url}
\usepackage[hidelinks]{hyperref}
\usepackage[utf8]{inputenc}
\usepackage[small]{caption}
\usepackage{graphicx}
\usepackage{amsmath}
\usepackage{array}
\usepackage{wrapfig}
\usepackage{multirow}
\usepackage{tabu}
\usepackage{booktabs}
\usepackage{subfigure}
\usepackage{dsfont,amsfonts,amssymb}

\title{GAP++: Learning to generate target-conditioned adversarial examples}

\author{
Xiaofeng Mao\footnote{Contact Author}\and
Yuefeng Chen\and
Yuhong Li\and
Yuan He\And
Hui Xue
\\
\affiliations
Alibaba Group\\
\emails
\{mxf164419, yuefeng.chenyf, daniel.lyh, heyuan.hy, hui.xueh \}@alibaba-inc.com
}

\begin{document}

\maketitle

\begin{abstract}
Adversarial examples are perturbed inputs which can cause a serious threat for machine learning models. Finding these perturbations is such a hard task that we can only use the iterative methods to traverse. For computational efficiency, recent works use adversarial generative networks to model the distribution of both the universal or image-dependent perturbations directly. However, these methods generate perturbations only rely on input images. In this work, we propose a more general-purpose framework which infers target-conditioned perturbations dependent on both input image and target label. Different from previous single-target attack models, our model can conduct target-conditioned attacks by learning the relations of attack target and the semantics in image. Using extensive experiments on the datasets of 
MNIST and CIFAR10, we show that our method achieves superior performance with single target attack models and obtains high fooling rates with small perturbation norms.
\end{abstract}

\section{Introduction}
Deep Neural Networks (DNNs) have been extensively studied and seen in most daily applications. However, recent works have demonstrated that DNNs are vulnerable to adversarial perturbations on various tasks, e.g., image recognition \cite{szegedy2013intriguing,goodfellow2014explaining}, object detection \cite{thys2019fooling} or semantic segmentation \cite{poursaeed2018generative}. These carefully crafted perturbations are imperceptible to humans but can mislead a well trained model after adding to the original inputs. 

Many attack algorithms have been proposed for generating such adversarial perturbations. Among them, model-based methods use feed-forward networks to generate adversarial perturbations. Prior to optimization-based methods, they need not to iteratively search the optimal perturbations in the pixel space, and improves the attack efficiency greatly. Additionally, model-based methods can produce adversarial examples without requiring access to the victim network once the attacker network is trained.

However, model-based methods are not practical enough currently for two reasons: 1) model training is a time-consuming process, and particularly, we need to retrain the model when the type of the attack (targeted or nontargeted) or the attack targets are changed. 2) adversarial examples generated by model-based methods have a poor transferability.

In this work, we aim at the first drawback mentioned above and propose an improved framework to generate target-conditioned perturbations.  Our method is motivated by \cite{8683008}, which uses conditional generative adversarial networks (cGAN) to generate the output conditioned on both given image and text. We think a good attacker should learn the relations of attack target and the semantics in image. Thus, it can generate not just the target-specific perturbations dependent on input images, but the image-specific perturbations conditioned on attack targets. For the latter, we named \textit{target-conditioned perturbations}. Different from previous methods which train one model for single target attack, our approach can generate all types of targeted perturbations using only one trained model. Furthermore, by treating the nontarget attack as a peculiar target attack, we use trained model for nontarget attack as well. Although the model deals with more attack tasks, the comparative experiment shows it achieves high fooling rates and superior performance with single target attack models.  

Our contributions can be summarized as follows:
\begin{itemize}
\item We present an improved model-based attack method which can generate target-conditioned perturbations under $L_{0}$, $L_{2}$ and $L_{\infty}$ norm. It greatly reduces the training cost and facilitates the attack process.
\item Our method can employ a trained targeted attack model for nontarget attack. It also achieves comparatively high fooling rates. 
\item The extensive experiments on 
MNIST and CIFAR10
illustrate that our method gains superior performance with single target attack models. 
\end{itemize}

\section{Related Work}
We introduce the recent work on adversarial examples and model-based attack in this section. 

\textbf{Adversarial Examples} Various techniques for crafting adversarial examples have been proposed recently. Traditional methods of adversarial attack can be divided into two classes: gradient-based approaches and optimization-based approaches. In gradient-based approaches such as the classic Fast Gradient Sign Method (FGSM) \cite{goodfellow2014explaining} and their targeted variants \cite{kurakin2016adversarial}, adversarial examples are generated by superimposing small perturbations along the gradient direction (with respect to the input image). FGSM uses the linear approximation of the gradient in deep networks whereas such models are commonly thought to be highly non-linear. It leads to sub-optimal results. To settle this problem,  \cite{kura2016adversarial,madry2017towards} iteratively take multiple small steps
while adjusting the gradient direction after each step and get better results. In optimization based approaches, adversarial perturbation is optimized for targeted attacks while satisfying certain constraints \cite{carlini2017towards,liu2016delving}. However, the optimization process is slow and can only optimize perturbation for one specific instance each time. Therefore, more and more works use feed-forward network to approximate optimal perturbations.

\textbf{Model-based attack methods}
Model-based attack methods use image generation \cite{goodfellow2014generative} or image-to-image translation \cite{isola2017image} techniques to create adversarial examples. This idea is first proposed by \cite{baluja2017adversarial}. The created adversarial examples can be universal \cite{hayes2018learning,poursaeed2018generative} or image-dependent \cite{xiao2018generating}. Model-based attack methods ensure the imperceptibility of perturbations in different ways. \cite{baluja2017adversarial} and \cite{xiao2018generating} minimize the $L_{2}$ distance between the original input and adversarial examples. \cite{li2019regional,poursaeed2018generative} embed normalization module in network to constrain the output directly. Another branch of the methods does not guarantee the imperceptibility of perturbations. They create adversarial examples closed in latent space instead of pixel space \cite{zhao2017generating}, or train a class-conditional GAN to generate adversarial examples in the same class \cite{tsai2018customizing,song2018constructing}. Thus, more unrestricted adversarial examples can be constructed in this case.





\begin{figure*}[htbp]
    \includegraphics[width=7in]{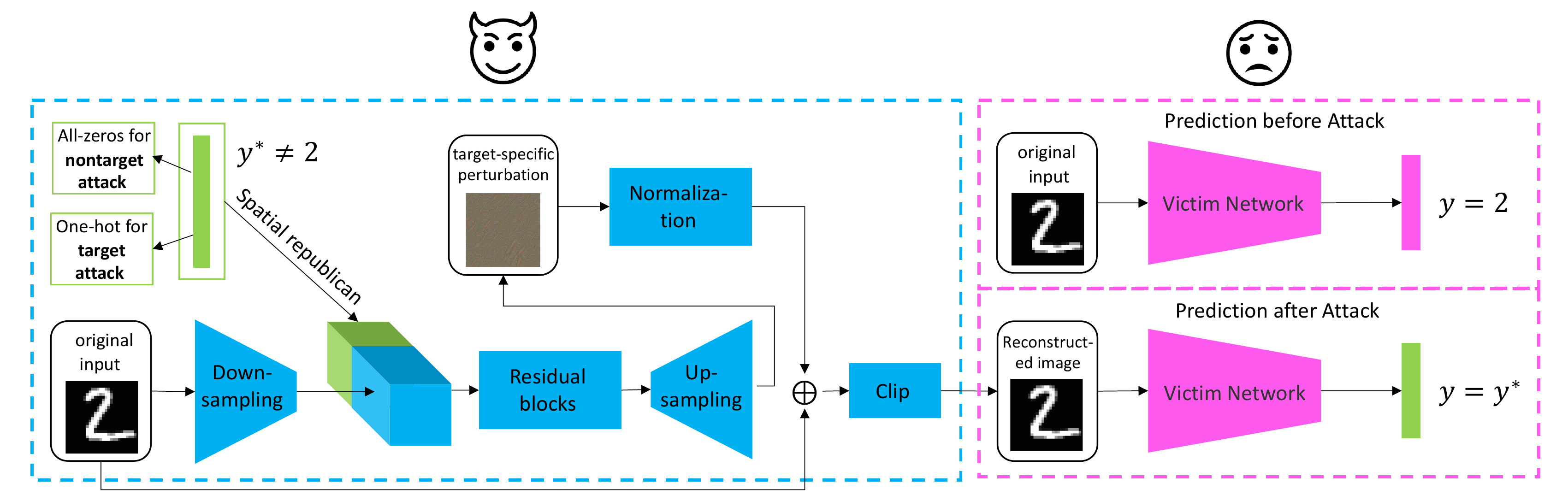}
    \caption{The overall architecture of our proposed method. Left part is the attacker network, and right part is the victim network (Best viewed in color). The parameters of victim network remain fixed during the training procedure. }
    \label{fig:archi}
\end{figure*}

\section{Proposed method}
Suppose $\mathcal{X}\subseteq \mathbb{R}^{n}$ be the pixel space, $\mathcal{Y}=\left \{ 1,2,...,C \right \}$ be the label space. Given original input $x\in \mathcal{X}$, a classifier $f(x):x\in \mathcal{X}\rightarrow y\in \mathcal{Y}$ and a target label $y^{\ast } \in \complement_{\mathcal{Y}} \left \{ f(x)\right \}$, the problem of generating targeted adversarial examples can be expressed as the optimization: $\mathrm{argmin}_{x^{\ast}}L(x,x^{\ast})\ s.t.\ f(x^{\ast})=y^{\ast}$, where
$L(\cdot )$ is a distance metric between examples from the input space (e.g., the L2 norm). Similarly, untargeted attack can also be defined as $\mathrm{argmin}_{x^{\ast}}L(x,x^{\ast})\ s.t.\ f(x^{\ast})\in \complement_{\mathcal{Y}} \left \{ f(x)\right \}$. We will go through the proposed scheme of both targeted and untargeted attacks in the rest of this section.

\subsection{Targeted attacks}
we firstly overview the general workflow of model-based adversarial attacks. The overall idea of model-based attacks is to train a neural network $g$ that transforms an input $x$ into an adversarial example $x^{\ast}$, which fools the classifier $f$ to output target label $y^{\ast}$. Formally, suppose that the neural network parameterized by $\theta$ is $g_{\theta}$, we have the following objective:
\begin{equation}
    \mathcal{L}_{targeted}=\sum_{x\in \mathcal{X}} \mathcal{H}(f(g_{\theta}(x)),y^{\ast})+L(x,g_{\theta}(x))
    \label{eq:1}
\end{equation}
where $\mathcal{H}$ is the cross-entropy loss. 

There are many improvements based on this framework. For example, GAP \cite{poursaeed2018generative} trains  $g_{\theta}$ to approximate the difference of input and adversarial images for better controlling over the perturbation magnitude. In other words, it generates adversarial examples $x^{\ast}$ by adding $\Delta x = g_{\theta}(x)$ to input $x$. In equation \ref{eq:1}, $L(\cdot)$ term makes the perturbation imperceptible. More advanced metrics such as soft hinge loss \cite{xiao2018generating}, perceptual similarity loss \cite{reddy2018nag} had been proposed. But optimizing  $L(\cdot)$ cannot guarantee that the model outputs adversarial examples which conform to the norm condition in the test phase. Therefore, another improvement is to normalize on the model output instead of adding an objective term in the training stage. Our work adopts these modification in the framework for getting preferable results.

Figure \ref{fig:archi} illustrates the overall architecture of the proposed model, which mainly consists of two parts: an attacker network $g$ and a victim network $f$. The attacker network takes the target label $y^{\ast}$ and original instance $x$ as its input. To add $y^{\ast}$ as conditional information, the attacker network can produce a distribution of perturbations, i.e.  $P(\Delta x|x,y^{\ast})$, which is conditioned on target labels. Then the output of the attacker is scaled to constrain its fixed norm. Specifically, if we have $\Delta x$, the normalized perturbations can be obtained by multiplying $\Delta x$ with $\min \left ( 1,\frac{\epsilon }{\left \| \Delta x \right \|_{p}} \right )$, where $\epsilon$ is the maximum permissible $L_{p}$ norm. Three types of normalization term are used in our model. $L_{\infty}$ and $L_{2}$ have been proposed in \cite{poursaeed2018generative}. Besides, we put forward the new $L_{0}$ norm for model-based attack. In our method, $L_{0}$ norm is implemented by simply selecting top-\textit{k} pixel values in $\Delta x$, where $k$ is given by users, similar to $\epsilon$. Next we add the normalized perturbations $\Delta x _{norm}$ and original input $x$ together, i.e. $x^{\ast} = \Delta x _{norm} + x$. In last stage, $x^{\ast}$ is clipped for ensuring it in the valid range of images and final result is the generated adversarial example.

Overall, the objective function of targeted attack is:
\begin{equation}
    \mathcal{L}_{targeted}=\sum_{x\in \mathcal{X}}\sum_{y^{\ast } \in \complement_{\mathcal{Y}} \left \{ f(x)\right \}} \mathcal{H}(f(g_{\theta}(x, y^{\ast})),y^{\ast})
    \label{eq:2}
\end{equation}
normalizing and clipping operations are omitted here for simplicity.

\subsection{Untargeted attacks}
In this section, we show how to employ our method to conduct untargeted attacks. Traditional methods using fooling loss to optimize attack algorithms. Fooling loss aims to maximize the cross-entropy between fooled prediction and original prediction:
\begin{equation}
    \mathcal{L}_{untargeted}=-\sum_{x\in \mathcal{X}} \mathcal{H}(f(g_{\theta}(x)),f(x))
    \label{eq:2}
\end{equation}
This maximize objective is incompatible with $\mathcal{L}_{targeted}$, which has a minimize objective. They cannot be optimized together. Alternatively, as proposed in \cite{kurakin2016adversarial} and \cite{kura2016adversarial}, we consider least-likely class $f_{ll}(x)=\arg \min f(x)$ as the target to train the model. In our architecture, every input image needs corresponding target label as condition information. However, no conditional target label exists in untargeted case. For this problem, as shown in Figure \ref{fig:archi}, we simply use zero vector for untargeted training. Because it does not affect the learning of internal representations by concatenating zero tensors in the model. The untargeted loss can be formulated as:
\begin{equation}
    \mathcal{L}_{untargeted}=\sum_{x\in \mathcal{X}} \mathcal{H}(f(g_{\theta}(x),\vec{0}),f_{ll}(x))
    \label{eq:2}
\end{equation}
By viewing least-likely untargeted attack as a special targeted attack. We adopt $\mathcal{L}_{untargeted}$ as the one training objective in the model. Now, we have the overall training objective as follows:
\begin{equation}
    \mathcal{L} = \mathcal{L}_{targeted} + \alpha \mathcal{L}_{untargeted}
\label{eq:2}
\end{equation}
where $\alpha$ is a balance coefficient. We set $\alpha$ to 1 as default in the rest of the paper.

Conventional model-based method needs to train models for attacks of different types or different targets. It leads to a great cost of time and computing resources. However, by using our approach, we can generate all types of adversarial examples just by training one model. In this aspect, our method is a lighter version which can be used for many attack tasks. Additionally, although the model deals with more attack tasks, our experiment shows it still has good attack performance.

\section{Experiments}
In this section, we first evaluate our model for both targeted and non-targeted white-box attack settings on the benchmarks of MNIST and CIFAR10. For each dataset, we simply use the training set to train the generative model and the validation set for evaluation. 


We will provide details of the network architecture and the training  procedure. The generation of pertubations has many shared properties with the image translation task, so our generative model mainly follows the architecture design of \cite{8683008}. Target labels as the conditioning infromation, are concatenated with the output of residual blocks before the deconvolution layers. The training time is partly determined by the class number of the recognition task. It takes half an hour to converge that training attack models on MNSIT and CIFAR10 task. 
The algorithm is runing on two Tesla P100 GPUs and optimized by Adam with the learning rate of 0.0002.  

\subsection{Experiment on MNIST and CIFAR10}
We first train three networks on the standard MNIST and CIFAR10 classification tasks. For MNSIT, we use 3-fully- connected-layer MLP network, ResNet18 and LeNet as the basic classifier. For more complicated CIFAR10 object recognition, three stronger model: SENet18, ResNet18 and VGG11 are used instead. Both the MNIST and CIFAR10 images are rescaled to 32$\times$32 size with 3 RGB channels for fitting the input size of the basic networks. To restrict the generated adversarial examples, we normalize the network output on different metrics. $L_{0}$, $L_{2}$ and $L_{inf}$ are used in the experiment (detailed analysis of different norm type can be seen in last paragraph of this section).

We compare the performance of our algorithm with FGSM and GAP. FGSM is the most representative gradient-based method which query the victim model at once and get the gradient of the loss function with respect to pixels, and moves a single step based on the sign of the gradient. GAP is a model-based method which use generative model to produce universal or image-dependent pertubations. Both of compared methods are fast and efficient. In order to ensure the fairness, we do not contrast with iterative optimized methods, such as Carlini and Wagner’s methods (CW). They attack slowly by querying many times to the target model, which are not practical in real-world scenario. 

\textbf{Performance of non-target attack.} We show the success rate of non-target attack in Table \ref{tabel:non-target}. All the methods are evaluated on the test partition using $L_{\infty}$ norm. As shown in the table, FGSM has lower attack success rate with others. The reason is that FGSM only uses a single direction based on the linear approximation of the gradient, which leads to sub-optimal
results. As a more powerful generative model, GAP get state-of-the-art attack success rate on both MNIST and CIFAR10 datasets. GAP only uses least-likely prediction as the desired class to produce the non-target adversarial samples. However, our approach fuses non-target and target attack tasks in one model. Surprisingly, although GAP++ deals with more tasks, the Table \ref{tabel:non-target} shows GAP++ has no performance loss but promotes the effect of non-target attacks slightly.
We analyze that the non-target attack can benefit from other tasks such as targeted attack in our model. The result suggests GAP++ surpass GAP 2\%-3\% on fooling rate averagely. 

Interestingly, we found the choice of the basic classification network will affect the success rate of attack in some cases. Specifically, for MNIST digit classification task, both MLP and ResNet18 is more vulnerable than LeNet. MLP has simpler network structure and less parameters, which leads to under-fitting problem. On the contrary, ResNet18 has more model capacity and may overfit on MNIST dataset. From this phenomenon, we analyse that any under-fitting or over-fitting will reduce robustness and increase vulnerability of the model. 

\begin{table}
\tiny
\centering
\renewcommand{\multirowsetup}{\centering}
\begin{tabular}{cccccccc}
\hline
\multirow{2}{3em}{Datasets}  & 
\multirow{2}{3em}{Attack type}
& \multirow{2}{3em}{Models} & \multirow{2}{3em}{AV\%} & 
\multirow{2}{3em}{AA\% $\epsilon$=0.05} & 
\multirow{2}{3em}{AA\% $\epsilon$=0.1} & 
\multirow{2}{3em}{AA\% $\epsilon$=0.15} & 
\multirow{2}{3em}{AA\% $\epsilon$=0.2} 
\\&&&&&&\\

\hline
\hline
\multirow{11}{3em}{MNIST}     &\multirow{3}{3em}{FGSM}  &    MLP &    97.5\%  &82.6\% &55.4\%&39.9\%&29.3\%\\
& & ResNet18  & 99.4\% &    85.7\% & 62.1\%&43.9\%&30.0\%\\ 
& & LeNet  & 99.2\% &    94.3\% & 85.1\%&68.3\%&43.1\%\\
&&&&&&&\\
&\multirow{3}{3em}{GAP}&MLP&97.5\%&\textbf{72.4\%}&44.7\%&\textbf{36.9\%}&28.0\%\\
& & ResNet18  &    99.4\% & 76.8\% & 58.3\%&40.5\%&30.7\%\\ 
& & LeNet  &    99.2\% &    85.2\% & \textbf{68.2\%}&57.5\%&46.1\%\\
&&&&&&&\\
&\multirow{3}{3em}{GAP++}&MLP&97.5\%&73.8\%&\textbf{43.3\%}&38.7\%&\textbf{27.0\%} \\
& & ResNet18  &    99.4\% &    \textbf{75.1\%} & \textbf{56.9\%}&\textbf{36.4\%}&\textbf{28.8\%}\\ 
& & LeNet  &    99.2\% &    \textbf{82.7\%} &68.5 \%&\textbf{55.4\%}&\textbf{40.8\%}\\
\hline
\multirow{11}{3em}{CIFAR10}     &\multirow{3}{3em}{FGSM}  &    SENet18 &90.5\%&88.3\%&80.0\%&78.9\%&72.2\%\\
& & ResNet18  & 92.9\% & 87.2\% &83.6\% &80.3\%&73.1\%\\ 
& & VGG11  & 89.3\% &    87.8\% & 83.6\%&82.1\%&73.5\%\\
&&&&&&&\\
&\multirow{3}{3em}{GAP}&SENet18&90.5\%&41.4\%&23.4\%&10.9\%&2.2\%\\
& & ResNet18  & 92.9\% & 47.5\% & 20.1\%&\textbf{5.4\%}&\textbf{0.0\%}\\ 
& & VGG11  & 89.3\% &  47.2\% & \textbf{21.1\%}&\textbf{7.8\%}&3.9\%\\
&&&&&&&\\
&\multirow{3}{3em}{GAP++}&SENet18&90.5\%&\textbf{33.7\%}&\textbf{18.0\%}&\textbf{6.2\%}&\textbf{0.0\%}\\
& & ResNet18  & 92.9\% & \textbf{44.3\%} & \textbf{18.0\%}&8.5\%&\textbf{0.0\%}\\ 
& & VGG11  & 89.3\% &  \textbf{41.9\%} & 21.3\%&9.2\%&\textbf{3.7\%}\\
\hline

\end{tabular}
\caption{Comparison of the non-targeted attack performance. We show the AV\% metric (Accuracy on the Validation set) and AA\% metric (Accuracy on the Adversary set) on three different attack methods: FGSM, GAP and GAP++. }
\label{tabel:non-target}
\end{table}

\begin{table}
\tiny
\centering
\renewcommand{\multirowsetup}{\centering}
\begin{tabular}{cccccccccccc}
\hline
\multirow{2}{3em}{Attack type}  & 
\multirow{2}{3em}{AA\% $\epsilon$=0.2}
& \multirow{2}{4em}{Target-1} & \multirow{2}{4em}{Target-2} & 
\multirow{2}{4em}{Target-3} & 
\multirow{2}{4em}{Target-4} & 
\multirow{2}{4em}{Target-5}
\\&&&&&&\\

\hline
\hline
FGSM  & 93.4\% & 11.4\%  &10.8\% &11.1\%&11.0\%&11.4\%\\
GAP-1&75.9\%&12.5\%&2.1\%&1.2\%&1.7\%&0.9\%\\
GAP-2&75.3\%&1.9\%&12.5\%&1.4\%&1.4\%&1.4\%\\
GAP-3&76.3\%&1.2\%&1.7\%&12.1\%&1.7\%&1.8\%\\
GAP-4&76.2\%&1.0\%&1.8\%&1.6\%&\textbf{12.2\%}&2.0\%\\
GAP-5&77.7\%&1.3\%&1.1\%&2.0\%&2.3\%&11.5\%\\
GAP++&50.5\%&\textbf{13.3\%}&\textbf{12.8\%}&\textbf{12.8\%}&12.0\%&\textbf{11.6\%}\\
\hline

\end{tabular}
\caption{Comparison of the targeted attack performance. We pick 5 different target labels to conduct the targeted attack on MNIST dataset using $\epsilon$=0.2 as the $L_{\infty}$ norm. GAP cannot generate various targeted perturbations, so we train the model for each target respectively. AA\% is calculated by averaging the accuracy on total 5 targeted adversaries.}
\label{tabel:target attack}
\end{table}

\textbf{Performance of target attack.} We make a comparative analysis of our method with FGSM and GAP. Table \ref{tabel:target attack} shows the experimental results, we use the accuracy of target class prediction as attack success rate. In the experiment, we choose 5 classes and calculate the prediction accuracy of each class. FGSM can generate arbitrary targeted attacks by calculating the gradient that reduces the loss of the target class, but it is not very accurate. Contrary to FGSM, GAP attacks target model with higher accuracy but it is unable to generate arbitrary targeted perturbations. So, for each target, we train a GAP model respectively (the model is tagged as GAP-1 if target of the attack is target-1). The result suggests each GAP model only focuses only on generating single target attacks. It signifies we must train 5 models to generate 5 types of target attacks. The proposed GAP++ overcomes this disadvantages of traditional model-based attack. By modeling the distribution of multi-targeted perturbations, we deal with all types of target attacks using one GAP++ model. Additionally, from the table \ref{tabel:target attack} we found even GAP++ generates more various perturbations, it gains superior performance with single targeted attack models. In terms of fooling rate, GAP++ achieves higher attack success rate than GAP because it modifies more samples to cheat the model. 
\begin{figure}[htbp]
    \subfigure[\tiny MNIST]{
    \includegraphics[width=1.63in]{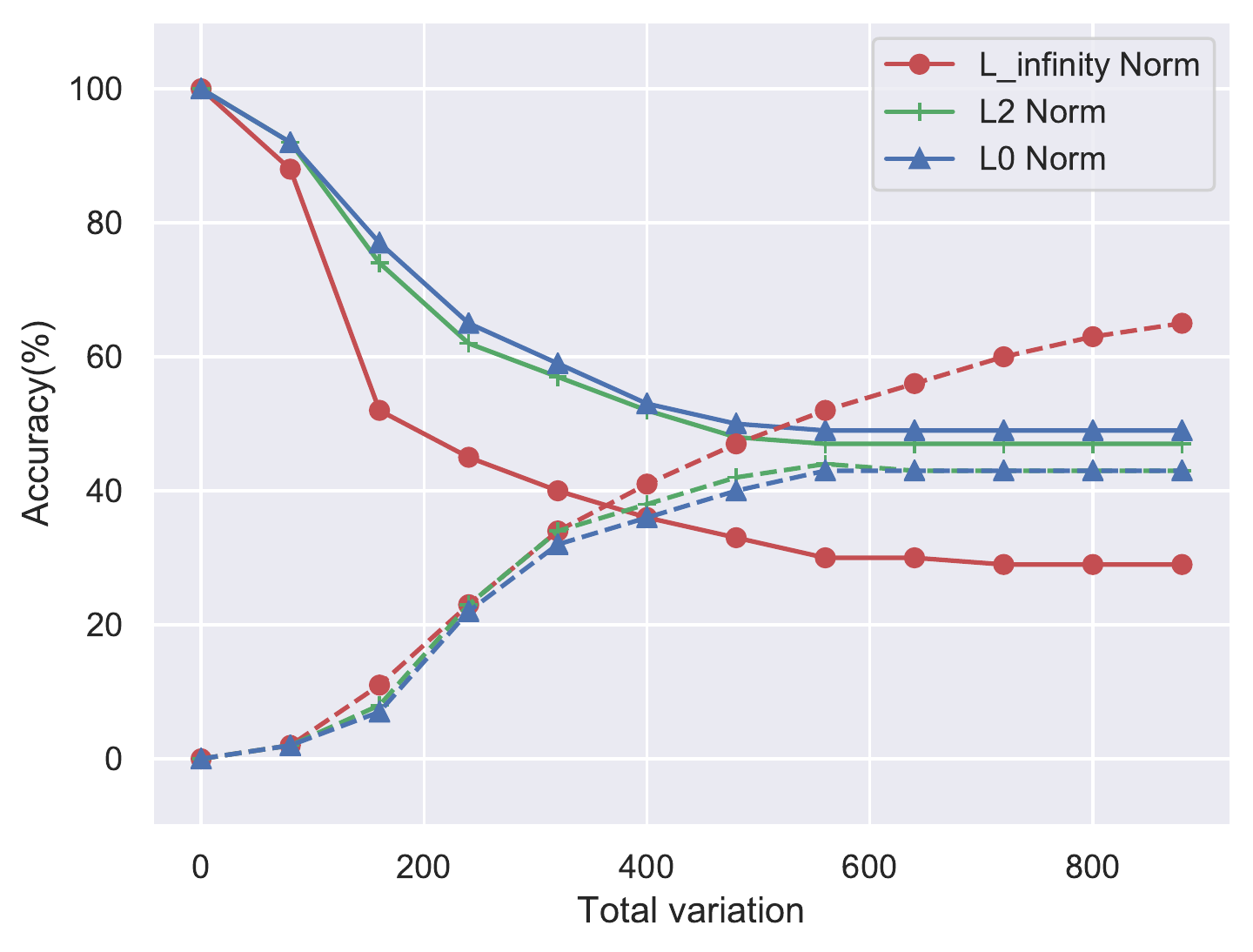}}
    \subfigure[\tiny CIFAR10]{
    \includegraphics[width=1.63in]{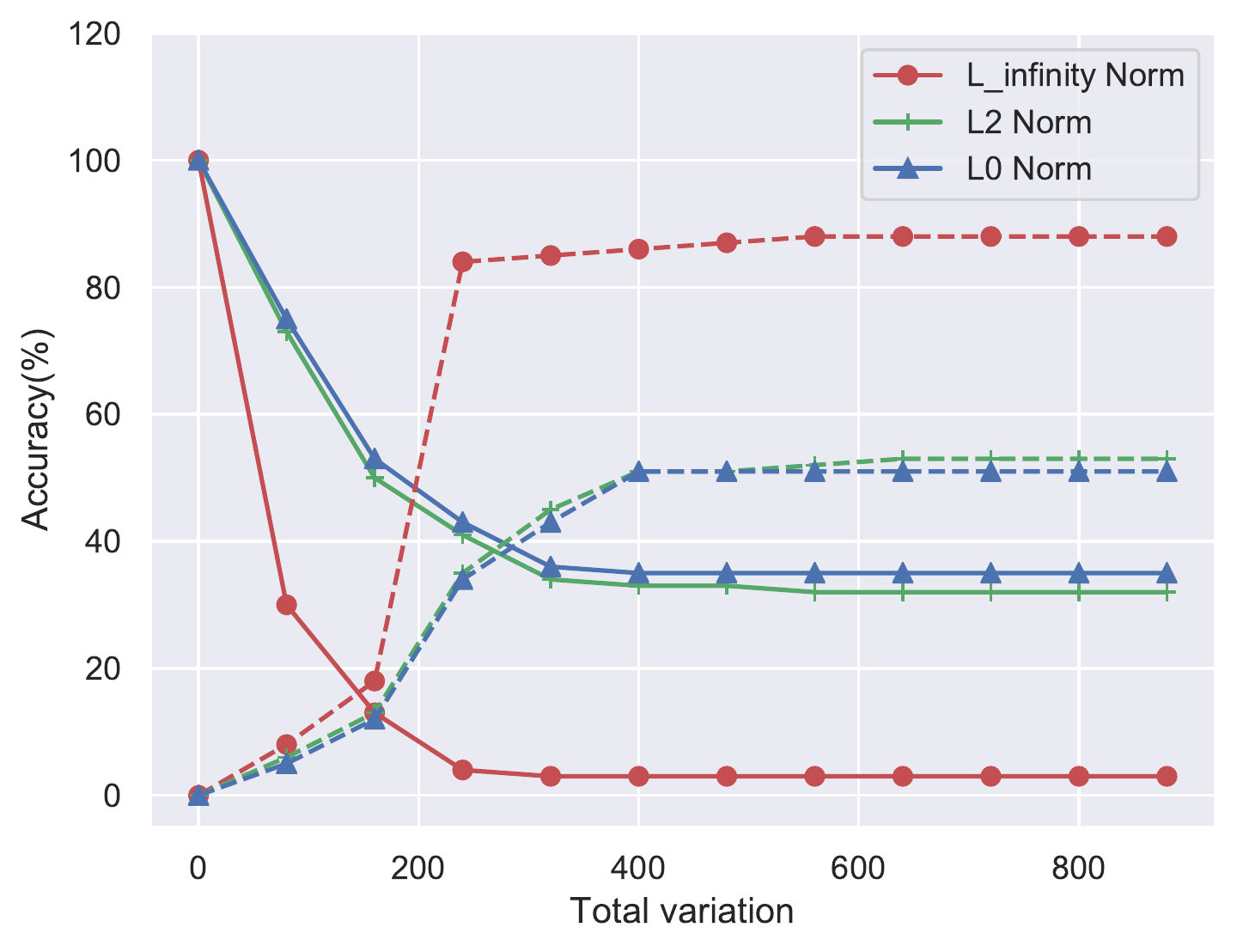}}
    \caption{The line chart compares the performance of GAP++ using different norm term: $L_{\infty}$, $L_{2}$ and $L_{0}$. We set the experiment under the targeted attack on ResNet18 model trained by MNIST and CIFAR10. The solid line indicates the model accuracy on generated adversarial samples, and the dotted line presents the targeted attack success rate. }
    \label{fig:norm}
\end{figure}

\begin{figure}[htbp]
\centering
    \includegraphics[width=3.4in]{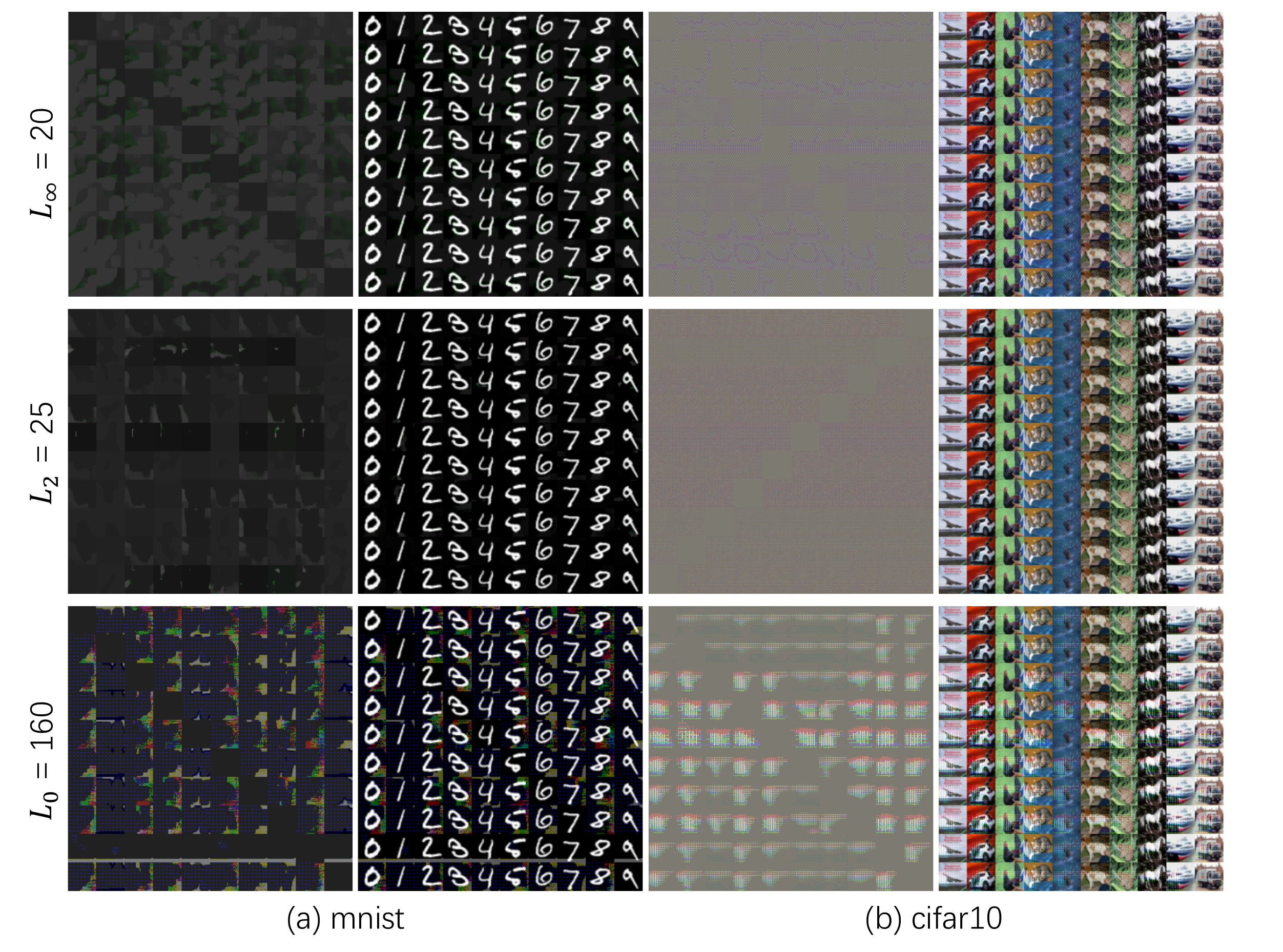}
    \caption{Visualization of targeted adversarial examples when total variation is 160 (corresponding $L_{\infty}=20$, $L_{2}=25$ and $L_{0}=160$). For each image grid, rows represent different targets and columns represent different original inputs.}
    \label{fig:visulize}
\end{figure}

\textbf{Norm constraint analysis.} We analyze three normalization term named $L_{\infty}$, $L_{2}$ and $L_{0}$ in GAP++. $L_{\infty}$ is a norm term which limits the maximum change of single pixel in adversarial samples. While $L_{2}$ constrains the sum of total pixel variations. $L_{0}$ introduces the sparsity to variations and aims to cheat the model by altering as few pixels as possible. We establish relation among these three norm terms by defining total variation metric, which represents the maximum variation of all pixels in the image for each norm term. As shown in Figure \ref{fig:norm}, $L_{\infty}$ achieves the best effect on both MNIST and CIFAR10 dataset, while $L_{2}$ and $L_{0}$ suffer a great performance loss. The result indicates $L_{2}$ and $L_{0}$ are still hard to applied on model-based adversarial attacks. By using $L_{\infty}$ norm, we found it is easy for model to learn the pattern or structure of perturbations instead of predicting the perturbation values directly. Although $L_{2}$ and $L_{0}$ normalize the output as well, however, the model tries hard to pick the right perturbation values from a wide range (0-128). Therefore, GAP++ trained on $L_{2}$ and $L_{0}$ norm can be more prone to fall into local optimum and output the sub-optimal results. We visualize the reconstructed adversary and its corresponding perturbations in Figure \ref{fig:visulize}. $L_{\infty}$ and $L_{2}$ norm makes the model to generate smooth perturbations. $L_{0}$ norm lets the model output salt-and-pepper perturbations, which can be clearly distinguished. From the perspective of perception, we can draw a conclusion that $L_{\infty}$ is a preferable normalization method.

\section{Conclusion}
In this paper, we propose a novel model-based adversarial attack method named GAP++, which can be used to generate target-conditioned  perturbations. GAP++ borrows the network architecture and normalization trick from original GAP, but achieves superior performance and lighter than GAP. Using extensive experiments on the datasets of MNIST and CIFAR10, we show that our method achieves high fooling rates with small perturbation norms. However, Our approach still behaves weakly in transferability, which is a problem to be solved in the future work.




\bibliographystyle{named}
\bibliography{ijcai19}

\begin{thebibliography}{}

\bibitem[\protect\citeauthoryear{Baluja and
  Fischer}{2017}]{baluja2017adversarial}
Shumeet Baluja and Ian Fischer.
\newblock Adversarial transformation networks: Learning to generate adversarial
  examples.
\newblock {\em arXiv preprint arXiv:1703.09387}, 2017.

\bibitem[\protect\citeauthoryear{Carlini and Wagner}{2017}]{carlini2017towards}
Nicholas Carlini and David Wagner.
\newblock Towards evaluating the robustness of neural networks.
\newblock In {\em 2017 IEEE Symposium on Security and Privacy (SP)}, pages
  39--57. IEEE, 2017.

\bibitem[\protect\citeauthoryear{Goodfellow \bgroup \em et al.\egroup
  }{2014a}]{goodfellow2014generative}
Ian Goodfellow, Jean Pouget-Abadie, Mehdi Mirza, Bing Xu, David Warde-Farley,
  Sherjil Ozair, Aaron Courville, and Yoshua Bengio.
\newblock Generative adversarial nets.
\newblock In {\em Advances in neural information processing systems}, pages
  2672--2680, 2014.

\bibitem[\protect\citeauthoryear{Goodfellow \bgroup \em et al.\egroup
  }{2014b}]{goodfellow2014explaining}
Ian~J Goodfellow, Jonathon Shlens, and Christian Szegedy.
\newblock Explaining and harnessing adversarial examples.
\newblock In {\em ICLR}, 2014.

\bibitem[\protect\citeauthoryear{Hayes and Danezis}{2018}]{hayes2018learning}
Jamie Hayes and George Danezis.
\newblock Learning universal adversarial perturbations with generative models.
\newblock In {\em 2018 IEEE Security and Privacy Workshops (SPW)}, pages
  43--49. IEEE, 2018.

\bibitem[\protect\citeauthoryear{Isola \bgroup \em et al.\egroup
  }{2017}]{isola2017image}
Phillip Isola, Jun-Yan Zhu, Tinghui Zhou, and Alexei~A Efros.
\newblock Image-to-image translation with conditional adversarial networks.
\newblock In {\em Proceedings of the IEEE conference on computer vision and
  pattern recognition}, pages 1125--1134, 2017.

\bibitem[\protect\citeauthoryear{Kurakin \bgroup \em et al.\egroup
  }{2016a}]{kura2016adversarial}
Alexey Kurakin, Ian Goodfellow, and Samy Bengio.
\newblock Adversarial examples in the physical world.
\newblock {\em arXiv preprint arXiv:1607.02533}, 2016.

\bibitem[\protect\citeauthoryear{Kurakin \bgroup \em et al.\egroup
  }{2016b}]{kurakin2016adversarial}
Alexey Kurakin, Ian Goodfellow, and Samy Bengio.
\newblock Adversarial machine learning at scale.
\newblock {\em arXiv preprint arXiv:1611.01236}, 2016.

\bibitem[\protect\citeauthoryear{Li \bgroup \em et al.\egroup
  }{2019}]{li2019regional}
Yingwei Li, Song Bai, Cihang Xie, Zhenyu Liao, Xiaohui Shen, and Alan~L Yuille.
\newblock Regional homogeneity: Towards learning transferable universal
  adversarial perturbations against defenses.
\newblock {\em arXiv preprint arXiv:1904.00979}, 2019.

\bibitem[\protect\citeauthoryear{Liu \bgroup \em et al.\egroup
  }{2016}]{liu2016delving}
Yanpei Liu, Xinyun Chen, Chang Liu, and Dawn Song.
\newblock Delving into transferable adversarial examples and black-box attacks.
\newblock In {\em ICLR}, 2016.

\bibitem[\protect\citeauthoryear{Madry \bgroup \em et al.\egroup
  }{2017}]{madry2017towards}
Aleksander Madry, Aleksandar Makelov, Ludwig Schmidt, Dimitris Tsipras, and
  Adrian Vladu.
\newblock Towards deep learning models resistant to adversarial attacks.
\newblock In {\em ICLR}, 2017.

\bibitem[\protect\citeauthoryear{{Mao} \bgroup \em et al.\egroup
  }{2019}]{8683008}
X.~{Mao}, Y.~{Chen}, Y.~{Li}, T.~{Xiong}, Y.~{He}, and H.~{Xue}.
\newblock Bilinear representation for language-based image editing using
  conditional generative adversarial networks.
\newblock In {\em ICASSP 2019 - 2019 IEEE International Conference on
  Acoustics, Speech and Signal Processing (ICASSP)}, pages 2047--2051, May
  2019.

\bibitem[\protect\citeauthoryear{Poursaeed \bgroup \em et al.\egroup
  }{2018}]{poursaeed2018generative}
Omid Poursaeed, Isay Katsman, Bicheng Gao, and Serge Belongie.
\newblock Generative adversarial perturbations.
\newblock In {\em Proceedings of the IEEE Conference on Computer Vision and
  Pattern Recognition}, pages 4422--4431, 2018.

\bibitem[\protect\citeauthoryear{Reddy~Mopuri \bgroup \em et al.\egroup
  }{2018}]{reddy2018nag}
Konda Reddy~Mopuri, Utkarsh Ojha, Utsav Garg, and R~Venkatesh~Babu.
\newblock Nag: Network for adversary generation.
\newblock In {\em Proceedings of the IEEE Conference on Computer Vision and
  Pattern Recognition}, pages 742--751, 2018.

\bibitem[\protect\citeauthoryear{Song \bgroup \em et al.\egroup
  }{2018}]{song2018constructing}
Yang Song, Rui Shu, Nate Kushman, and Stefano Ermon.
\newblock Constructing unrestricted adversarial examples with generative
  models.
\newblock In {\em Advances in Neural Information Processing Systems}, pages
  8312--8323, 2018.

\bibitem[\protect\citeauthoryear{Szegedy \bgroup \em et al.\egroup
  }{2013}]{szegedy2013intriguing}
Christian Szegedy, Wojciech Zaremba, Ilya Sutskever, Joan Bruna, Dumitru Erhan,
  Ian Goodfellow, and Rob Fergus.
\newblock Intriguing properties of neural networks.
\newblock In {\em ICLR}, 2013.

\bibitem[\protect\citeauthoryear{Thys \bgroup \em et al.\egroup
  }{2019}]{thys2019fooling}
Simen Thys, Wiebe Van~Ranst, and Toon Goedem{\'e}.
\newblock Fooling automated surveillance cameras: adversarial patches to attack
  person detection.
\newblock {\em arXiv preprint arXiv:1904.08653}, 2019.

\bibitem[\protect\citeauthoryear{Tsai}{2018}]{tsai2018customizing}
Shih-hong Tsai.
\newblock Customizing an adversarial example generator with class-conditional
  gans.
\newblock {\em arXiv preprint arXiv:1806.10496}, 2018.

\bibitem[\protect\citeauthoryear{Xiao \bgroup \em et al.\egroup
  }{2018}]{xiao2018generating}
Chaowei Xiao, Bo~Li, Jun-Yan Zhu, Warren He, Mingyan Liu, and Dawn Song.
\newblock Generating adversarial examples with adversarial networks.
\newblock In {\em Proceedings of the 27th International Joint Conference on
  Artificial Intelligence}, pages 3905--3911. AAAI Press, 2018.

\bibitem[\protect\citeauthoryear{Zhao \bgroup \em et al.\egroup
  }{2017}]{zhao2017generating}
Zhengli Zhao, Dheeru Dua, and Sameer Singh.
\newblock Generating natural adversarial examples.
\newblock {\em arXiv preprint arXiv:1710.11342}, 2017.

\end{thebibliography}
\end{document}